\documentclass[preprint,journal]{IEEEtran}

\usepackage{cite}
\ifCLASSINFOpdf
\usepackage[pdftex]{graphicx}
\graphicspath{{./Figures/}}
\DeclareGraphicsExtensions{.pdf,.jpeg,.png}
\else
\fi

\usepackage{color}

\usepackage{amssymb}

\usepackage[cmex10]{amsmath}
\usepackage{algorithmic}

\usepackage{algorithm}

\usepackage[hidelinks]{hyperref}

\usepackage{url}

\usepackage{tikz}

\renewcommand{\thepage}{DOI: 10.1109/TASE.2015.2432746 ~\arabic{page}}

\begin{document}
	
	 \onecolumn 
	 \textbf{	 © 2016 IEEE. Personal use of this material is permitted. Permission from IEEE must be
	 	obtained for all other uses, in any current or future media, including
	 	reprinting/republishing this material for advertising or promotional purposes, creating new
	 	collective works, for resale or redistribution to servers or lists, or reuse of any copyrighted
	 	component of this work in other works.}		

\textbf{--\\
		\\
		DOI:\url{10.1109/TASE.2015.2432746}}

	\newpage
	 \twocolumn 
%
\title{An Iterative Approach for Collision Free Routing and Scheduling in Multirobot Stations}

\author{Domenico~Spensieri,
        Johan~S.~Carlson,
        Fredrik~Ekstedt,
        and~Robert~Bohlin
\thanks{All authors are with the Geometry and Motion Planning Group, at the Fraunhofer-Chalmers Research Centre for Industrial Mathematics, Gothenburg, 41288, Sweden, e-mail: domenico.spensieri@fcc.chalmers.se}
}

\maketitle


\begin{abstract}
This work is inspired by the problem of planning sequences of operations, as welding, in car manufacturing stations where multiple industrial robots cooperate. The goal is to minimize the station cycle time, \emph{i.e.} the time it takes for the last robot to finish its cycle. This is done by dispatching the tasks among the robots, and by routing and scheduling the robots in a collision-free way, such that they perform all predefined tasks. We propose an iterative and decoupled approach in order to cope with the high complexity of the problem. First, collisions among robots are neglected, leading to a min-max Multiple Generalized Traveling Salesman Problem (MGTSP). Then, when the sets of robot loads have been obtained and fixed, we sequence and schedule their tasks, with the aim to avoid conflicts. The first problem (min-max MGTSP) is solved by an exact branch and bound method, where different lower bounds are presented by combining the solutions of a min-max set partitioning problem and of a Generalized Traveling Salesman Problem (GTSP). The second problem is approached by assuming that robots move synchronously: a novel transformation of this synchronous problem into a GTSP is presented. Eventually, in order to provide complete robot solutions, we include path planning functionalities, allowing the robots to avoid collisions with the static environment and among themselves. These steps are iterated until a satisfying solution is obtained. Experimental results are shown for both problems and for their combination. We even show the results of the iterative method, applied to an industrial test case adapted from a stud welding station in a car manufacturing line.
\\
\\
\emph{Note to Practitioners}---This article is motivated by the problem of planning robot operations in welding applications in the automotive industry. Here, a number of welding tasks have been introduced along the car body: the goal is to let the robots perform such tasks while minimizing the cycle time (or \emph{makespan}). The main difficulties, from the manufacturing engineer perspective, lie in assigning the tasks to the robots, deciding the order and the timing of the operations, avoiding collisions between the robots and the environment, and among the robots themselves. We present in this work an iterative approach, consisting of two steps: first, sequences for the robot operations are computed in order to minimize the cycle time, while neglecting collisions among robots; then, given the assignment of tasks to robots, the operations are reordered and scheduled while avoiding conflicts among robots. Robot motions are also automatically computed to avoid collisions with the static environment. We show an optimal algorithm, for the first part, based on implicit enumeration (branch and bound) and introduce a novel suboptimal algorithm, for the second part, to synchronize the robots. These algorithms are iterated while fetching information about the problem that are hard to compute, thus following a lazy approach. Tests on problems adapted from the literature and from the automotive industry, show clear improvements over more sequential approaches and good running times. The algorithms are especially suited for cases with up to 40 tasks and 4 robots, as for typical \emph{geometry stations}. In future works, we will further investigate efficient heuristic optimization approaches in order to handle larger problems, consisting of more than 100 tasks, as for typical \emph{assembly stations}.

\end{abstract}



\IEEEpeerreviewmaketitle


\section{Introduction}

\IEEEPARstart{I}{n} the last decades, for the automotive manufacturers short production times, on one hand, and large volumes, on the other, have become crucial. In order to achieve that, efficient equipment utilization is needed. Nevertheless, resource optimization plays even an important role when considering sustainable production systems, both economically and for the ecological aspects, in terms of less energy consumption and working space, see \cite{SUST_EQUIP}.

\begin{figure}
\centering
\includegraphics[width=2.5in]{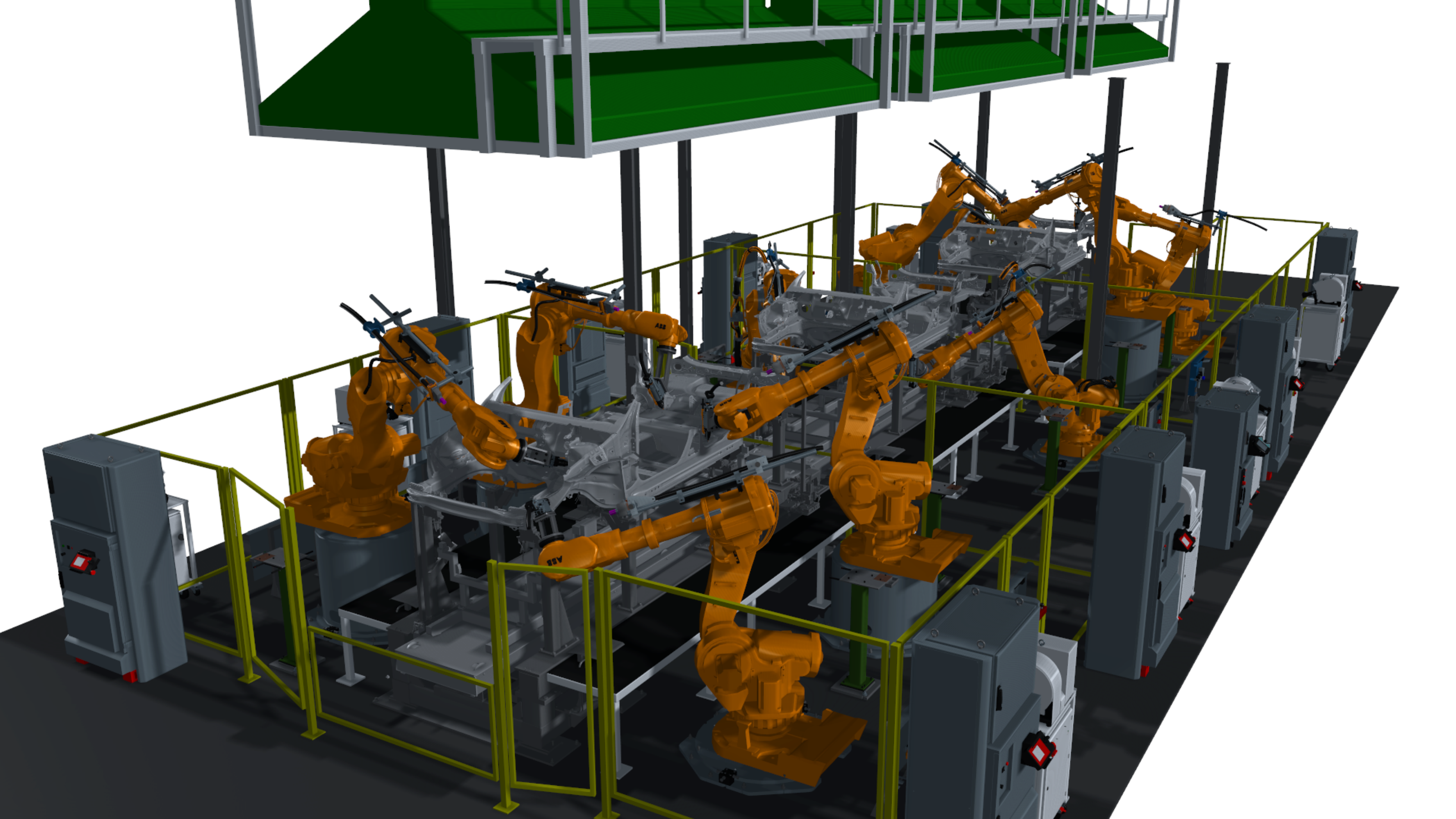}
\caption{A car body assembly line equipped with stud welding robots, modeled in the simulation software IPS, Industrial Path Solutions, \cite{IPS-online}.}
\label{fig_stud_line}
\end{figure}

In this work, we restrict the focus to the automotive manufacturing process. The car body, also called "Body in White" (BiW), is usually assembled together on a line where several stations are placed serially and in parallel. At each station multiple robots perform operations like stud, spot welding, and sealing, and often share the same workspace, see Fig. \ref{fig_stud_line}. 

In this framework we are mainly concerned with the problem of maximizing the number of products assembled in a line. This can be directly translated into minimizing the time needed at each station to perform predefined tasks in a collision-free way. If a manufacturing line reduces the cycle time by $r$ ($0<r<1$), then it would potentially increase the number of assembled products by $r/(1-r)$. For a car manufacturer, this strong relation means that reducing the cycle time, for example, by 33\% would directly translate into 50\% more cars produced. The immediate impact is tremendous.

A very powerful way to shorten ramp up times and increase the volumes is to simulate as soon as possible all the upcoming processes, including the design and the manufacturing ones. 
Simulation software is therefore strongly needed and provides the main key to virtual manufacturing. Unfortunately, nowadays, there is still much work done manually inside the software tools, which is both time consuming and error prone. On the other side, few or sometimes no automatic software tools at all are available, due to the difficulties in modeling the process and to the complexity of synthesizing solutions.

Here, we describe an algorithm and an automatic tool, aiming at planning robot motions in order to achieve near-optimal programs, performing all predefined tasks in the station, avoiding collisions between them and the environment, and among the robots themselves.

\section{Problem description and related literature}

The challenge is to concurrently: 

\begin{itemize}
\item distribute the tasks among robots such that each of them is assigned to one robot (set partitioning, dispatching),
\item decide in which way each robot should perform a task among several alternatives,
\item find a sequence of tasks for each robot (routing), 
\item compute robot paths that are collision-free w.r.t the static environment (path planning), 
\item schedule the sequences of paths such that no collision occurs among the robots (scheduling).
\end{itemize}

The problem in its entire complexity has received attention only in the last five years, at the best of the authors' knowledge.

A reason for that might be that it is very interdisciplinary, integrating together combinatorial optimization and path planning issues. Moreover, only in the last years, with the development of computers and software tools, it has been possible to face large problems in a reasonable time. Anyway, a wide amount of literature has been produced in each of these two fields. There are several works regarding the Vehicle Routing Problem (VRP), which is a more general case of the Multiple Traveling Salesman Problem (MTSP); for a good review of the VRP and its variations see \cite{TOTH}.
Furthermore, several articles have dealt with path planning for mobile and industrial robots; for a good treatment of the subject see \cite{LATOMBE_BOOK} and \cite{LAVALLE_BOOK}. Here, we refer to the works that most closely relate to ours, combining together the sub-problems described above.

From a combinatorial optimization perspective, if we disregard collisions among robots, the problem presented in this work can be modeled as a min-max Generalized Multiple Traveling Salesman Problem (min-max MGTSP), or as an uncapacitated min-max Generalized Vehicle Routing Problem, (min-max GVRP), with multidepots. In contrast with the classical VRP, in the generalized one, GVRP, customers are clustered in groups, and it is enough to visit one customer in each group. The other difference is that the objective to minimize is the length of the longest tour (min-max), not the sum of the tours lengths (min-sum), as in the more classical formulation.

Little work has been produced in this area, compared to the wide literature referring to the min-sum MTSP and VRP.

An exact algorithm for the min-max VRP, able to prove the optimality of a solution provided to the problem presented at the Whizzkids '96 competition, has been proposed by Applegate \emph{et al.} in \cite{APPLEGATE}. The algorithm is based on branch-and-cut and requires a very lare computing time on a computer network. Some heuristic approaches have also been investigated, see \cite{NA}. In this work, the author also adapts into No Depot min-max MTSP, and solves some instances from the TSPLIB, see \cite{TSPLIB}, thus giving results to publicly available benchmark instances. In \cite{JOHN_CARLS} a load balancing algorithm for min-max VRP is provided. It runs by iterating the solution of a linear programming sub-problem and of a TSP, to achieve the minimization of the longest tour, without explicitly considering the min-max as objective function. There, theoretical lower bounds are provided for problem instances in the 2D Euclidean space, based on geometrical space partitions. The GVRP with min-sum objective has also been studied. Ghiani \emph{et al.} in \cite{GHIANI} introduced an efficient transformation into a Capacitated Arc Routing Problem (CARP). Anyway, we are mostly interested in the min-max GVRP, where there is a lack of work.

All the articles so far cited do not deal at all with resource allocation problems such as conflicts between the tours. By conflict we mean that parts of the tours may not occur at the same time. These types of conflict are often introduced as a way of modeling physical collisions among moving objects, \emph{e.g.} industrial/mobile robots or ground/air vehicles. Only in the last years the problem has been treated. In \cite{EEK} a min-max MTSP with conflicts is solved by using a genetic algorithm and local search heuristics. In \cite{SPENS} the problem is generalized to a min-max MGTSP with conflicts and solved exploiting genetic algorithm as well, by using the solution of a GTSP as local search. Two works trying to exploit the geometrical properties of the problems are \cite{SEGEB1} and \cite{SEGEB2}. Here, the assignment of welding tasks to robots exploits their geometrical distribution, in order to achieve a good trade-off between: separating the tasks among robots to avoid collisions vs. partitioning them to balance the robot loads and minimize the longest tour. The method is applied to a complete line composed of three stations with a total of ten robots. In \cite{WELZ}, the Welding Cell Problem (WCP) is introduced, which is a variant of the min-max MTSP with conflicts. Here, the problem is to find routes for robots such that the tour time for each robot is within a predefined maximum time. The authors discuss also the min-max MTSP. The approach uses a branch and bound (B\&B) method together with column generation. A related problem, the Laser Source Problem, is studied in \cite{RAMB1}. It is similar to the min-max MTSP with conflicts, but with an additional requirement, which is to find the minimum number of laser sources. The motivation is the same, a multirobot station for car body manufacturing, and their solving approach is to use NP-hard sub-problems in a B\&B framework. Conflicts are handled as well and results for problem instances with up to 40 jobs are illustrated. High computing times are however needed for large instances and no path planning capability to avoid conflicts is included.

Another similar problem is the one of routing of automated guided vehicles (AGVs) and supervision of automated manufacturing systems (AMSs), where a number of unmanned vehicles operate in a factory, or warehouse environment transporting goods between various places along a network of uni- or bi-directional lanes. A good overview of concepts and methods for AGV routing and scheduling is provided by \cite{QUI} and \cite{GANES}. The main differences between our problem and the AGV scheduling and routing, in general, are the absence, in our settings, of a predefined roadmap network and the presence of several alternatives that perform the tasks. 

A large amount of literature concerning AGVs and AMSs is related to discrete event modeling and different extensions of Petri nets are used, see \cite{WU_ZENG,ZHOU_1,HU_1,HU_3,HU_4,HU_5,HU_6,WU_ZHOU_LI,LI_ZHOU_WU,LI_WU_ZHOU}, for example. In these works, often, the main focus is on modeling and on synthesizing a maximally permissive supervisor or to enforce and guarantee \emph{liveness} of the system. Liveness is a property implying that no \emph{deadlock} occurs. Deadlock is a very separated concept from the one of \emph{conflict} used here. A conflict occurs when two agents or jobs want to use the same resource or machine: it is explicitly specified within the problem definition. A deadlock is a state requiring four conditions: mutual exclusion, no preemption, hold while wait and circular wait. It is not necessarily identified at the problem definition level, since its detection often requires deep analysis of the system. Note that there exist problems such as the classical job shop scheduling problem (JSSP) that
\begin{itemize}
\item do contain conflicts, defined by the fact that two operations may not be done at the same time on the same machine, and
\item do not generate any deadlock in their open-loop behavior since the "hold while wait" condition is not present as stated in \cite{HUANG}.
\end{itemize}

The papers \cite{FANTI,FANTI_ZHOU} also take a look at alternative graph-based approaches.

In \cite{ZHOU_2,NISHI_ANDO_KONISHI,NISHI_MAENO}, various optimization is present, but with different objectives and/or conditions than in our case. In the first case, the problem consists in routing AGVs with only one fixed start and one fixed goal in a predefined network. In the second case, robot energy is minimized with collision avoidance as a side-constraints, and in the third case total transport time is minimized.

The conflict-free routing problem, given a fixed scheduling, is addressed in \cite{KIM} utilizing a Dijkstra-like procedure on a time-window graph. In \cite{KRIS}, a column generation method is used to minimize the makespan given a fixed set of assigned jobs. In \cite{REVEL} a more dynamic approach for conflict resolution is suggested, computing AGV routes incrementally and thereby enabling conflict avoidance under changing circumstances using a modified Banker's algorithm. In \cite{NISHI} the problem is decomposed into a master problem for the task assignment (dispatching) and for routing, and a sub-problem for vehicle routing. It is solved by subdividing the time horizon into time slots and applying MIP algorithms. The network on which vehicles can move is precomputed. This characteristic is also present in \cite{CORREA} and \cite{KHAYAT} where Constraint Programming and MIP techniques are used. In \cite{KHAYAT}, the assignment of operations to machines is already determined together with the order in which operations will be done by several jobs. The problem is very close to job shop scheduling. In \cite{CORREA}, results for different types of problems show that for \emph{compact or spread out} requests sets there is a natural decoupling between the scheduling and routing problems. 

That property, together with the geometrical characteristics about the problems addressed here, motivate the strategy to consider the two extremes of a scale. On the one hand, we try to generate lower bounds by assuming that no collision between the robot paths is present. To do that, we use a B\&B approach, naturally branching on the tasks to be assigned, as in \cite{RAMB1}, but we generalize it to solve the min-max MGTSP. Moreover, we present two different ways to improve the lower bounds, thus decreasing the number of expanded nodes and running times. On the other hand, we want to generate feasible solutions even for cluttered environments, where a lot of collisions may occur, by preferring as quality measure the feasibility of the solution to the cycle time minimization. To achieve that, we introduce here a novel algorithm that solves a specialized synchronous routing and scheduling problem, in which the assignment of tasks to robots is fixed and robots are assumed to move synchronously. The algorithm proposed is based on a new transformation of the problem into a GTSP. This may be interpreted as an attempt to redesign robot paths by means of combinatorial optimization. In the rest of this article we will use the terms \emph{robots} and \emph{agents} indifferently.


\section{Graph formulation and notation}

The problem can be represented in graph terms. Given $N_A$ agents $a_1, a_2,...,a_{N_A}$ and a set $G=\{g_1,\ldots,g_{N_A},\ldots,g_{N_G}\}$ of $N_G$ number of tasks, all possible ways that agent $a_k$ can perform the tasks in $G$ may be represented by a set $V^k$. Each agent has also a home position, which may be achieved by several configurations. We have included these home tasks into the set of tasks $G$ and denoted them $g_1,\ldots,g_{N_A}$, without loss of generality. Thus, each $V^k$ can be partitioned into groups (sometimes called clusters in the literature) and the information about which elements of $V^k$ perform a specific task is given by the function $\gamma^k:V^k\rightarrow G$, which maps each element into a task in $G$. In the rest of the paper we will denote with $D^k_i$ the set of vertices in $V^k$ that can perform task $g_i$, \emph{i.e.} $D^k_i=\{v\in V^k:\gamma^k(v)=g_i\}$. 
A set of edges $E^k \subseteq V^k \times V^k$ can be naturally added to the set of vertices, thus forming the graph $(V^k,E^k)$. This graph can be extended into a weighted one by the help of a non-negative weight (or cost) function $c^k:E^k\rightarrow \mathbb{R^+}$. Since we assume this cost function to be symmetric, we will talk about arcs, instead of edges. The weight function models the time needed for agent $a_k$ to move from a start configuration $u\in V^k$ to an end configuration $v\in V^k$, with $u\neq v$, by $c^k(u,v)$. As a special case the time needed to perform task $\gamma^k(v)$ by configuration $v$ is modeled as $c^k(v,v)$: we will, in the rest of the paper, indicate this processing time with $c^k(v)$ and omit the $k$ since the $c^k(u, v)$ is completely determined by the vertices $u, v$.
At this point we may define a tour $T^k$ for agent $a_k$ as a sequence of vertices, starting and ending at a vertex belonging to the corresponding home task $g_k$: $T^k=\{u^k_1,\ldots,u^k_{N_{T^k}},u^k_1\}$, where $u^k_i\in V^k, \forall i=1,\ldots,N_{T^k}$, and such that $\gamma^k(u^k_i) \neq \gamma^k(u^k_j), \forall i\not=j$ and $\gamma^k(u^k_1) = g_k$.
The cost associated to a tour $T^k$ is given by

\begin{equation}
\label{eq_TOUR_DEF}
c(T^k)=\sum \limits_{i=1}^{N_{T^k}-1} \left[ c(u^k_i)+c(u^k_i,u^k_{i+1}) \right] + c(u^k_{N_T^k}) + c(u_{N_{T^k}},u^k_1)
\end{equation}

Given a subset of tasks $\tilde G \subseteq G$, one can write the induced GTSP as

\begin{subequations}
\label{GTSP_DEF}
minimize $c(T^k)$ \emph{s.t.}
\begin{align}
\forall g \in \tilde G, \exists u \in T^k:\gamma^k(u)=g
\end{align}
\end{subequations}

The best value obtained is refered as $GTSP(\tilde G)$.
A feasible solution to the min-max MGTSP is an ordered set of tours  $T=(T^1,\ldots,T^{N_A})$, such that the corresponding sets $(G^1,\ldots,G^{N_A})$ form a partition of $G$, \emph{i.e.} $G^k \cap G^l = \emptyset, \forall k\neq l$ and $\bigcup\limits_{k=1}^{N_A} G^k=G$. The cost associated to $T $ is given by

\begin{equation}
\label{GMTSP_DEF}
c(T)=\max \limits_{k=1,\ldots,N_A} \{c(T^k)\}
\end{equation}

The min-max MGTSP with \emph{conflicts} has to take also care of incompatible arcs between the tours. A \emph{conflict} between two arcs  $(s^k_i, s^k_j)$ and $(s^l_m, s^l_n)$ exists if they are defined to be incompatible by the modeled process. This is often introduced in order to model geometrical collisions between robot paths or paths where the distance between the robots is under the minimum allowed clearance threshold.

We can define a scheduled tour as a tour with time information, that is a re-parametrization of the tour itself
\begin{equation}
\label{eq_SCHEDULED_TOUR}
T^k_{scheduled} = \left(t[s^k_1]^L, \ldots, t[s^k_i]^A, t[s^k_i]^L, \ldots, t[s^k_1]^A \right)
\end{equation}
where,  $t[s^k_i]^A$ is the time agent arrives at $s^k_i$ and $t[s^k_i]^L$ is the time the agent leaves the same point. Given scheduled tours, two conflicting arcs $(s^k_i, s^k_{i+1})$ in $T^k$ and $(s^l_m, s^l_{m+1})$ in $T^l$, with $k \neq l$, define an \emph{active conflict} if $ \left[ t[s^k_i]^L, t[s^k_{i+1}]^A \right] \cap \left[ t[s^l_m]^L, t[s^l_{m+1}]^A \right] \neq \emptyset $, \emph{i.e.} if they overlap in time. At the same way, there is an active conflict between a vertex $s^k_i$ and arc $(s^l_m, s^l_{m+1})$ if $ \left[ t[s^k_i]^A, t[s^k_i]^L \right] \cap \left[ t[s^l_m]^L, t[s^l_{m+1}]^A \right] \neq \emptyset $, and between vertices $s^k_i$ and $s^l_m$ if $ \left[ t[s^k_i]^A, t[s^k_i]^L \right] \cap \left[ t[s^l_m]^A, t[s^l_m]^L \right] \neq \emptyset $. Note that when we identify the cost with time, then we have $c(T^k) = t[s^k_1]^A$.

An optimal solution to the min-max MGTSP with conflicts will consist in an ordered set of scheduled tours, with minimum makespan, such that no active conflict is present.


\section{Optimal solution for min-max MGTSP without conflicts}

Achieving optimality in absence of conflicts is very important in terms of cycle time, because the solution obtained:

\begin{itemize} 
\item can result to be collision-free due to the geometrical properties of the problems;
\item can be a very good initial guess for improvement-based heuristic algorithms;
\item can be used to dispatch the tasks among the agents and then resolving conflicts by avoiding the possibility to redistribute tasks, with the aim to decrease complexity;
\item can be velocity tuned to avoid conflicts and still be of very high quality;
\item from a practical perspective, can be modified to be collision-free by small manual adjustments.
\end{itemize} 

Therefore, in order to find these potential best solutions, we relax the problem by assuming that there are no conflicts among the agents arcs. This relaxation transforms the whole problem into a min-max MGTSP, \emph{i.e} without conflicts. The problem instances faced in this work contain a small number of tasks, under 40, and can well model some real world applications like \emph{geometry stations}. For larger problems, more sophisticated algorithms have to be provided. 

In this section, we discuss an optimal algorithm and show how this can be easily implemented. One practical advantage is the ease of implementation, due to the absence of bounds based on LP (Linear Programming) formulations, which often require sophisticated implementations exploiting sparsity and dealing with ill conditioned instances.

\subsection{Branch and bound based algorithm}
\label{sec_BB}

The general branch and bound (B\&B) architecture in this work follows the lazy pattern described in \cite{CLAUSEN}. Each sub-problem consists of a GTSP: they are solved to optimality, due to the limited size of the instances, by a dynamic programming approach, which is a generalization of \cite{HELD_KARP}.

Along the search in the B\&B tree, at each node, we maintain two sets, $G_S$ and $G_N$ that partition $G$. $G_S$ contains the tasks assigned to some of the agents, whereas the remaining tasks, the ones not yet assigned to anyone, are the elements of $G_N$. $G_S$ can be further partitioned into $N_A$ subsets indicated by $G^k_S$, with $k=1,\ldots,N_A$: each set $G^k_S$ represents the tasks in $G_S$ assigned to agent $a_k$.

The \emph{selection} of the next node to be expanded in the B\&B may be done by depth-first, breadth-first, best-first strategies or by some other criterion. Our experience has shown that depth-first performs well in terms of computing times and has the enormous advantage of keeping the dimension of the nodes queue limited. This avoids state space explosion.

The \emph{branching} step consists in assigning, if possible, an element $g_i \in G_N$ to each of the agents, thus creating at most $N_A$ children nodes. The group to be assigned can be randomly chosen in $G_N$, or decided based on some heuristics that exploit information specific to the problem.

\begin{algorithm}
\caption{Creating children nodes in the branching step of B\&B}
\label{alg_BRANCHING}
\begin{algorithmic}
\STATE{$G_N \leftarrow G_N \setminus \{g_i\}$}
\FOR{$k=1$ to $N_A$}
\IF{$D^k_i \neq \emptyset$}
\STATE{NEWNODE($G^1_S,\ldots,G^k_S \cup \{g_i\},\ldots,G^{N_A}_S, G_N$)}
\ENDIF
\ENDFOR
\end{algorithmic}
\end{algorithm}

The bounding step computes a \emph{lower bound} for the current node. The first part of the bound is solving $N_A$ GTSPs, one for each agent. The lower bound $\underline{b}$ is then
\begin{eqnarray} 
\label{eq_LOWER_BOUND}
\underline{b}^k(G^k_S) =& GTSP(G^k_S) & \forall k=1,\ldots,N_A \label{eq_LB}\\
\underline{b} =& \max \limits_{k=1,\ldots,N_A} \{\underline{b}^k(G^k_S)\} \nonumber
\end{eqnarray}
Note that if the triangle inequality holds among the vertices of the problem, then it is possible to guarantee that the value obtained is actually a lower bound on the optimal value, see Section \ref{theo_TRIANG_INEQ}.

When reaching a leaf in the tree, \emph{i.e.} when $G_S= \bigcup \limits_{k=1}^{N_A} G^k_S = G$, the bound obtained is equal to an \emph{upper bound} $\overline{b}$ for the entire problem, \emph{i.e.} the min-max MGTSP, without conflicts. If the bound $\overline{b}$ computed is less than the best bound $\overline{b}_{MIN}$ so far found, we set $\overline{b}_{MIN}=\overline{b}$. This method will be referred as Approach 1.

Anyway, we can further improve the algorithm performance by using additional knowledge about the process. Indeed, modeling operations like welding times, on one hand, and modeling robot reachability constraints, on the other, can be utilized to improve the lower bounds.

\subsection{Tightening the bound}

 Tasks may consist in closing, welding and opening a gun, sealing or other operations. The time it takes for the accomplishment of one such task is modeled to be constant, namely $c_G$. This is due to the fact that, often, the actual duration of such operations is considered to not heavily vary among different robots and operations. The engineers responsible for the simulations insert this value as input to the algorithms described in this article. This characteristic may be exploited to improve the branching step and to tighten the lower bounds. To do that, we use the following lemma:

\newtheorem{theorem}{Lemma}
\begin{theorem}
\label{theo_TRIANG_INEQ}
Given a best tour $T^*_n$ with length $c(T^*_n)$ for an SGTSP $G_n$ with $n$ groups, and an SGTSP $G_{n+1} = G_n \cup \{g_{n+1}\}$, if the triangle inequality holds, then
\begin{equation} 
\label{theo_TRIANGLE_INEQ}
c(T^*_{n+1}) \geq \max \{c(T^*_n) + 2 \min (g_{n+1}, G_n) - \max (G_n), c(T^*_n)\}
\end{equation}
where $c(T^*_{n+1})$ is the length of an optimal tour in $G_{n+1}$,
\begin{eqnarray} 
\label{theo_TRIANG_INEQ_EXPLAIN}
\min (g_{n+1}, G_n) & = & \min \limits_{\substack{p_i \in G_n \\ p_k \in g_{n+1}}} c(p_i, p_k) \nonumber\\
\max (G_n) & = & \max \limits_{\substack{p_i \neq p_j \in G_n}} c(p_i, p_j)\nonumber
\end{eqnarray}

\end{theorem}
For a proof of the above Lemma, see Appendix \ref{APPENDIX}. The idea with the first part is to avoid generating B\&B nodes where vertices far away from small clusters are added. The second part of the inequality, $c(T_{n+1}^*) \geq c(T_n^*)$, is just a motivation for why the B\&B is consistent.

\begin{algorithm}
\caption{Revised branching step exploiting geometrical information}
\label{alg_TRIANGLE_INEQ}
\begin{algorithmic}
\STATE{$G_N \leftarrow G_N \setminus \{g_i\}$}
\FOR{$k=1$ to $N_A$}
\STATE{$\delta = 2\min (g_i, G^k_S) - \max(G^k_S$)}
\IF{$D^k_i \neq \emptyset$ \AND $\max \{\underline{b}^k(G^k_S), \underline{b}^k(G^k_S) + \delta\}$ $<\overline{b}_{MIN}$}
\STATE{NEWNODE($G^1_S,\ldots,G^k_S \cup \{g_i\},\ldots,G^{N_A}_S, G_N$)}
\ENDIF
\ENDFOR
\end{algorithmic}
\end{algorithm}

Nonetheless, also the computation of the lower bound can be improved. In fact, at each node in the B\&B tree, the remaining tasks, \emph{i.e.} the ones represented as elements of  $G_N$, have to be performed eventually by some agent. In the classical formulation of the GVRP and of the MGTSP, all vertices can be reached by all agents (vehicles or salesmen). In a typical multirobot station, however, this is not always true. Tasks, for example, may be outside the working space of some robots or collisions are not avoidable.

That leads to the following model, where the values obtained by the solution of the GTSPs are increased with the minimal value it will take to cover all remaining tasks.

\begin{subequations} \label{GENERAL_SET_PART}
minimize $c$ \emph{s.t.}
\begin{align}
c - c_G\sum\limits_{i:g_i \in G_N} x_{ik} &\ge \underline{b}^{k}(G^k_S) & \forall k=1,\ldots,N_A \label{GENERAL_SET_PART_1}\\
\sum\limits_{k=1}^{N_A} x_{ik} &= 1 & \forall i:g_i \in G_N \label{GENERAL_SET_PART_2}\\
x_{ik} &= 0 & \forall i,k:D^k_i=\emptyset \label{GENERAL_SET_PART_3}\\
x_{ik} \in &\{0,1\} & \forall i:g_i \in G_N \nonumber\\
 & & \forall k=1,\ldots,N_A \nonumber
\end{align}
\end{subequations}

The cycle time is indicated with $c$. The unknowns $x_{ik}$ take the value 1 if task $g_i$ is assigned to agent $a_k$, take the value 0 otherwise. Constraints (\ref{GENERAL_SET_PART_1}) state that the cycle time should be greater than or equal to the sum of the best tour among the groups in $G^k_S$, plus the sum of the processing times for the task assigned to agent $a_k$. Equalities (\ref{GENERAL_SET_PART_2}) establish that each task should be performed by exactly one agent. Constraints (\ref{GENERAL_SET_PART_3}) model that task $g_i$ is unreachable for agent $a_k$.

Problem \ref{GENERAL_SET_PART} can be solved by general purpose Mixed Integer Linear Programming (MILP) packages. This method will be referred as Approach 2.

Anyway, in some cases, these packages do not return any optimal solution in a reasonable time. Therefore, we relax problem (\ref{GENERAL_SET_PART}) by eliminating (\ref{GENERAL_SET_PART_3}), meaning that all agents can perform all tasks. The new problem becomes:

\begin{subequations} 
\label{GREEDY_SET_PART}
minimize $c$ \emph{s.t.}
\begin{align}
c - c_G n_k &\ge \underline{b}^{k}(G^k_S) & \forall k=1,\ldots,N_A\\
\sum\limits_{k=1}^{N_A} n_k &= |G_N| & \\
n_{k} \in & \mathbb{N} & \forall k=1,\ldots,N_A \nonumber
\end{align}
\end{subequations}

In this formulation the unknowns $n_k$ are the number of tasks assigned to agent $a_k$, which relates to the variables in (\ref{GENERAL_SET_PART}) by $n_k = \sum\limits_{i:g_i \in G_N} x_{ik}$. The problem can be optimally solved in polynomial time by a greedy algorithm, \emph{i.e.} by assigning one task at the time to the agent with the shortest tour. For a closed form expression, see Appendix \ref{APPENDIX}. This method will be referred as Approach 3.

\subsection{Simulation Results}
\begin{center}
\begin{table*}[ht]
\renewcommand{\arraystretch}{1.3}
\caption{Comparison of B\&B properties among different lower bounds used, on geometric instances from TSPLIB with four agents.}
\label{table_comparison}
\centering
\begin{tabular}{| c|c || c|c || c|c || c|c |}
\hline
 & & Approach 1 &  & Approach 2  & & Approach 3 & \\
\hline
Problem name & Best solution & N. nodes & Time (s) & N. nodes & Time (s) & N. nodes & Time (s)\\
\hline
11eil51 		&	 72.804 		&	 60 			&	0.000		&	 47 		&	 0.015 		&	46		&	0.093	\\
\hline															
14st70 			&	 190.036 		&	 425 			&	0.015		&	 360 		&	 0.016 		&	345		&	0.468	\\
\hline															
16eil76			&	 131.724 		&	 2912 			&	0.047		&	 2410 		&	 0.047 		&	2325	&	6.334	\\
\hline															
16pr76 			&	 34001.008 		&	 360 			&	0.031		&	 263 		&	 0.031 		&	258		&	0.515	\\
\hline															
20kroA100 		&	 7038.138 		&	 2925 			&	0.094		&	 2281 		&	 0.094 		&	2137	&	6.115	\\
\hline															
20kroB100 		&	 6545.117 		&	 3533 			&	0.156		&	 2950 		&	 0.156 		&	2471	&	4.977	\\
\hline															
20kroC100 		&	 5071.016 		&	 4499 			&	0.124		&	 3208 		&	 0.094 		&	2679	&	12.792	\\
\hline															
20kroD100 		&	 6357.273 		&	 12919 			&	0.250		&	 9718 		&	 0.249 		&	9022	&	18.720	\\
\hline															
20kroE100 		&	 6751.198 		&	 2263 			&	0.078		&	 1816 		&	 0.078 		&	1766	&	5.273	\\
\hline															
20rat99			&	 292.655 		&	 7866 			&	0.187		&	 6048 		&	 0.172 		&	5476	&	13.759	\\
\hline															
21eil101 		&	 178.327 		&	 317570 		&	4.524		&	 266014 	&	 4.352 		&	231098	&	476.115	\\
\hline															
21lin105 		&	 4622.421 		&	 7960 			&	0.203		&	 6499 		&	 0.219 		&	6372	&	12.698	\\
\hline															
22pr107 		&	 15698.472 		&	 219 			&	0.078		&	 110 		&	 0.047 		&	110		&	0.172	\\
\hline															
25pr124 		&	 22462.762 		&	 23355 			&	0.530		&	 16677 		&	 0.468 		&	15022	&	24.898	\\
\hline															
26bier127 		&	 32507.758 		&	 6885 			&	0.297		&	 4144 		&	 0.265 		&	3749	&	14.196	\\
\hline															
26ch130 		&	 1578.800 		&	 26701 			&	0.561		&	 10799 		&	 0.343		&	10615	&	32.448	\\
\hline															
28pr136 		&	 25694.316 		&	 109615 		&	2.886		&	 48242 		&	 1.622 		&	47049	&	257.511	\\
\hline															
29pr144 		&	 22378.426 		&	 148788 		&	3.557		&	 103607 	&	 3.339 		&	103024	&	95.129	\\
\hline															
30kroA150 		&	 7166.586 		&	 153316 		&	4.181		&	 79230 		&	 3.151 		&	77698	&	609.933	\\
\hline															
30kroB150 		&	 7309.708 		&	 425300 		&	11.809		&	 241628 	&	 7.051 		&	232518	&	1223.001	\\
\hline															
31pr152			&	 28024.425 		&	 29786 			&	6.177		&	 19071 		&	 3.479 		&	17303	&	55.318	\\
\hline															
32u159 			&	 13413.603 		&	 1607531 		&	31.544		&	 715318 	&	 17.301 	&	706262	&	1796.913	\\
\hline															
39rat195 		&	 645.101 		&	49179533		&	2184.513	&	 4535693 	&	 335.761 	&	4298503	&	44344.205	\\
\hline															
40d198			&	 4779.423 		&	1076643			&	1946.830	&	 385042 	&	 824.200	&	371681	&	1599.696	\\
\hline															
40kroA200 		&	 8106.633 		&	27993063		&	1012.104	&	 3860702 	&	 185.283	&	3820629	&	24813.831	\\
\hline
\end{tabular}
\end{table*}
\end{center}

Here, we show some results from running this algorithm on modified SGTSP instances from the TSPLIB, see \cite{TSPLIB}, adapted to our problem. When it comes to the MILP part to solve (\ref{GENERAL_SET_PART}), we have used the COIN-OR package, \cite{COINOR}.

Table \ref{table_comparison} shows the results of the computations on geometrical instances of the problem, based on the Euclidean distance. The table compares 3 different bounds:

\begin{enumerate}
\item approach 1 uses (\ref{eq_LOWER_BOUND}) to bound the function;
\item approach 2 improves it by using Lemma \ref{theo_TRIANG_INEQ}, Algorithm \ref{alg_TRIANGLE_INEQ}, and the solution to problem (\ref{GENERAL_SET_PART}); 
\item approach 3 replaces the solution to problem (\ref{GENERAL_SET_PART}) with the one to problem (\ref{GREEDY_SET_PART});
\end{enumerate} 

As expected, the number of expanded nodes decreases from approach 1, towards approach 3 and approach 2, and so almost do the running times. Note that, for the largest instance, computing times can almost be 10 times faster than with the first approach.

Problem (\ref{GENERAL_SET_PART}) has too large memory requirements when solved by COIN-OR, thus we use it only if the number of tasks is less than 10 (using approach 3 otherwise). Even in that case, there are large computing times that do not justify its use with respect to the greedy bound computed in Approach 3.

We have also run simulations where the processing time $c_G$ at each task was decreased. We have noticed that, by decreasing $c_G$, the advantages of approach 2 are not as evident, whereas the bound computed as in \ref{alg_TRIANGLE_INEQ} gives some little improvements.

Experiments by changing the initial solution have also proven its enormous effect on the dimension of the explored state space.

These characteristics motivate, on the one side, the study for faster solution to problem (\ref{GENERAL_SET_PART}), and, on the other, the use of other neighborhoods and meta-heuristics for the search, to improve initial solutions.

At the end of the B\&B, an optimal solution with value $\overline{b}_{MIN}$ is obtained, consisting of a tour for each agent $T^k=(s^k_1, \ldots,s^k_{N_{T_k}}, s^k_1)$, where the tasks have been partitioned into $(G^1,\ldots,G^{N_A})$, with $G^k = \bigcup \limits_{i=1}^{N_{T^k}} s^k_i$.


\section{Synchronous routing and scheduling}
\label{sec_SRS}

Here, we fix the load for each agent to $(G^1,\ldots,G^{N_A})$ and let each $G^k$ be composed of exactly one element per group. This partition may be the result of the B\&B algorithm above, in which case each $G^k=\{s^k_1, \ldots,s^k_{N_{T^k}}\}$, or is due to a process specific fixed assignment. On the other hand, reordering the groups within one agent, and tuning the velocities are allowed. The goal is now to find a sequence of vertices for each agent such that no collision occurs.

In order to cope with the high complexity of the problem, we introduce the constraint that the agents move synchronously. This means that they will move from one task to another, starting and finishing simultaneously. As special case one or more agents can remain still, while others might move. This assumption also finds concrete applications such as the one in \cite{AKELLA}, and in the latest controllers family from some robot manufacturer, see ABB MultiMove \cite{ABB}.

A natural way of modeling the problem is by using a synchronized state space $S_G$, consisting of the Cartesian product of the sets $G^k$, \emph{i.e.} $S_G = G^1 \times \ldots \times G^{N_A}$.

This problem is not very common in the literature. The closest work found is \cite{AKELLA}. There, inspired by laser drilling applications for microelectronics manufacturing system, the authors solve the coverage of planar points by multiple robots, that is routing and scheduling with collision avoidance constraints. Since the application has thousands of points, the problem is divided into two sub-problems: a so called splitting problem and an ordering problem, which are solved by heuristics.

When dropping the synchronous assumption, the problem arisen is the so called laser sharing problem, see \cite{RAMB2}, where the assignment of jobs to robots is fixed, and is indicated as "RSP-J" in the same paper. The problem is solved by a B\&B that generates a large number of sequences.

The solution obtained in this way can always be improved by relaxing back to the original problem, fixing the tours, and using a velocity tuning algorithm, see Section \ref{sec_asynchronous}. We propose here solving this problem by creating a GTSP. The original problem is re-modeled with the aim to apply a known solving algorithm for the transformed problem.

Each group in the transformed model will be visited once, but the original vertices might be reached more than once: this can facilitate the solution of some problems.


\subsection{Two agents case}
\label{sec_SYNCH_2_AGENTS}

We now want to find a sequence of states in $S_G = G^1 \times G^2$ $= \{ (s^1_1,s^2_1),\ldots, (s^1_m,s^2_1),\ldots, (s^1_m,s^2_n),\ldots,(s^1_{N_{T_1}},s^2_{N_{T_2}}) \}$, such that, when projecting a state onto $G^1$, respectively  $G^2$, all their elements are visited. In other words, the goal is to visit all tasks modeled as element of $G^1$, resp. $G^2$.
\begin{eqnarray} 
\label{eq_GROUPS}
G^1 = \{ s_1^1, s_2^1,\ldots, s_i^1,\ldots, s_m^1 \ldots, s^1_{N_{T_1}} \} \nonumber\\
G^2 = \{ s_1^2, s_2^2,\ldots, s_j^2,\ldots, s_n^2,\ldots, s^2_{N_{T_2}} \}
\end{eqnarray}
Each state $s \in S_G$ can be uniquely identified by its two components $s_i$ in $G^1$ and $s_j$ in $G^2$, and be referred as $s_{ij}$:
\begin{equation} 
\label{eq_CART_PROD}
S_G = \{ s_{11}, \ldots, s_{i1},\ldots, s_{ij},\ldots,s_{N_{T_1} N_{T_2}} \}
\end{equation}
The set $S_G$ is then provided with a set of arcs between its vertices, thus forming a graph $(S_G, S_G \times S_G)$. Since the agents move synchronously, the cost (time) for the arc $(s_{ij}, s_{mn})$ is given by

\begin{equation}
\label{L_INF_NORM}
c(s_{ij}, s_{mn}) = \max \{c(s^1_i,s^1_m), c(s^2_j,s^2_n)\}
\end{equation}

The objective is to find a sequence that minimizes cycle time while avoiding conflicting states.

\begin{figure}
\centering
\def\svgwidth{2.5in}
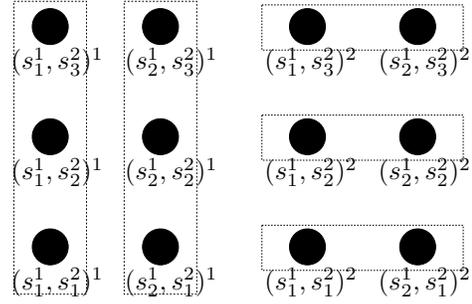
\caption{The synchronized state space with its clone, and the "vertical" and the "horizontal" groups.}
\label{fig_doublestatespace}
\end{figure}

In order to obtain such a sequence, we present here a new transformation into a well known problem, the GTSP, for which efficient algorithms are known. The solution to the GTSP, then, can be always transformed back into a solution to the original problem. To achieve that we generate a second layer $S_G^2$ that is a copy of the first one, $S_G$ (from now on referred as $S_G^1$). The vertices are grouped together in the following way to create a GTSP: vertices in $S_G^1$ are clustered in "vertical" groups, one for each column, whereas vertices in $S_G^2$ are clustered in "horizontal" groups, one for each row, see Fig. \ref{fig_doublestatespace}. More formally, the first layer consists of $N_{T^1}$ groups $F_i, \forall i = 1,\ldots,N_{T^1}$

\begin{equation}
\label{VERTICAL_GROUPS}
F_i = \{ (s^1_i, s^2_j)^1, \forall j=1,\ldots,N_{T_2} \}
\end{equation}

The second layer will have $N_{T^2}$ groups $F_{j+N_{T^1}}, \forall j = 1,\ldots,N_{T^2}$
\begin{equation}
\label{HORIZONTAL_GROUPS}
F_{j+N_{T^1}} = \{ (s^1_i, s^2_j)^2, \forall i=1,\ldots,N_{T^1} \}
\end{equation}
The idea behind this grouping is to create a problem aiming at fulfilling the constraints of visiting each original vertex at least once. Then, to minimize cycle time avoiding collisions, arcs are added between these vertices, and weighted according to (\ref{L_INF_NORM})
\begin{multline}
\label{L_INF_NORM_LAYERS}
c\left ( (s^1_i,s^2_m)^{l_1}, (s^1_j,s^2_n)^{l_2} \right ) = c(s^{l_1}_{ij}, s^{l_2}_{mn}) =\\
= c(s_{ij}, s_{mn}), \forall l_1, l_2 \in \{1,2\}
\end{multline}
If there is a conflict between these vertices, then the cost will be $\infty$. Thus, a GTSP has been defined consisting of $2N_{T_1}N_{T_2}$ vertices and $N_{T_1} + N_{T_2}$ groups.
\\
Resuming: one can create a GTSP by defining 

\begin{itemize}

\item a set of vertices $S^1_G \cup S^2_G =$
\begin{equation}
\label{eq_VERTICES}
\{ s^1_{11},\ldots, s^1_{mn},\ldots,s^1_{N_{T_1} N_{T_2}}, s^2_{11},\ldots, s^2_{mn},\ldots,s^2_{N_{T_1} N_{T_2}} \}
\end{equation}
\item the distances among them:
\begin{multline}
\label{eq_VERTICES}
c(s^1_{ij}, s^1_{mn}) = c(s^1_{ij}, s^2_{mn}) = c(s^2_{ij}, s^1_{mn}) = c(s^2_{ij}, s^2_{mn}) = \\
= \max \{c(s^1_i,s^1_m), c(s^2_j,s^2_n)\}
\end{multline}
\item the organization of vertices into groups:
\begin{eqnarray} 
\label{eq_GROUPS}
F_1 & = & \{ s^1_{1j}, 						\forall j=1,\ldots,N_{T_2} \} \nonumber\\
F_2 & = & \{ s^1_{2j}, 						\forall j=1,\ldots,N_{T_2} \} \nonumber\\
\ldots \nonumber\\
F_{N_{T^1}} & = & \{ s^1_{N_{T^1}j}, 		\forall j=1,\ldots,N_{T_2} \} \nonumber\\
F_{1+N_{T^1}} & = & \{ s^2_{i1}, 			\forall i=1,\ldots,N_{T^1} \} \nonumber\\
\ldots \nonumber\\
F_{N_{T^2}+N_{T^1}} & = & \{ s^2_{iN_{T^2}},\forall i=1,\ldots,N_{T^1} \} \nonumber\\
\end{eqnarray}
\end{itemize}

Note that each group $F_i$ is visited exactly once, but vertices in $G_1$ and $G_2$ may occur more than once. This makes the problem closer to a path planning problem, where agents search for a collision-free way to move. Indeed, by allowing that, one may obtain better solutions for instances modeling very cluttered environments. This is the main advantage of this algorithm against other meta-heuristic search algorithms, which explore a more limited state space. Note that a predefined roadmap where agents can move on is not present here as for AGV applications.

\begin{figure}
\centering
\def\svgwidth{2.5in}
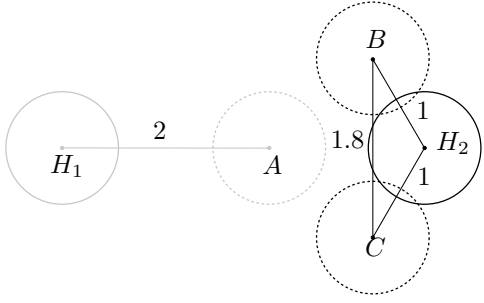
\caption{An example with two agents where the transformation to a GTSP finds a better solution by re-routing and scheduling the agents: geometrical view.}
\label{fig_rerouting}
\end{figure}

\begin{figure}
\centering
\def\svgwidth{2.5in}
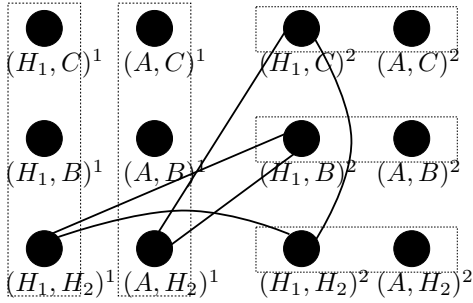
\caption{An optimal solution to the problem represented in Fig. \ref{fig_rerouting}.}
\label{fig_ReroutingStateSpace}
\end{figure}

Consider, \emph{e.g.} Fig. \ref{fig_rerouting}: there are two agents with circular shape at their home position $H_1$, and $H_2$ respectively. After the B\&B phase they are assigned, respectively vertex $(A)$, and $(B, C)$, \emph{i.e.} $G_1 = \{H_1, A\}$ and $G_2 = \{H_2, B, C\}$. By a standard approach, since they collide when moving from $H_1$ to $A$, resp. from $B$ to $C$, the best cycle time will be 6.8. At the contrary, by transforming the problem into a GTSP, one obtains the solution in Fig. \ref{fig_ReroutingStateSpace}, which has a total cost of 6: $(H_1, H_2)$, $(H_1, B)$, $(A, H_2)$, $(H_1, C)$, $(H_1, H_2)$. Note that the problem has been transformed into a symmetric one, therefore also the reversed sequence is an optimal one.

The advantage of this transformation is that a vertex is allowed to be visited more than once, allowing to obtain shorter tours in some cases. However, some problems remain still unsolvable, even if a feasible solution exists, see Appendix \ref{APPENDIX_2}. Note that shorter tours may be obtained because the arcs in $S_G \times S_G$ do not satisfy the triangle inequality. They are not the result of an optimal path planning algorithm that exploits all the degrees of freedom of both robots simultaneously.

\subsection{N agents case}
\label{sec_SYNCH_N_AGENTS}

The generalization to the case of $N_A$ agents is straightforward and can be done by building $N_A$ layers. Each layer $S_G^k$ consists of $|S_G| = \prod\limits_{k=1}^{N_A} N_{T^k}$ vertices, grouped in $N_{T^k}$ groups, in the following way. Group $F^k_i$ in the $k$-th layer will be:
\begin{eqnarray} 
\label{eq_VERTICAL_GROUPS}
F^k_i = & \{ (s^1_{j_1}, \ldots, s^k_{j_k}, \ldots, s^{N_A}_{j_{N_A}})	& : j_k = k,\nonumber\\
		&																&j_i = 1, \ldots, N_{T_k},\\
 		& 																& \forall i = 1, \ldots, N_A, i \neq k\} \nonumber\\
 		& 																& \forall k=1,\ldots,N_A \nonumber
\end{eqnarray}
Thus, the GTSP built has a total of $N_A |S^G|$ vertices distributed within $\sum\limits_{k=1}^{N_A} N_{T^k}$ groups. Note that the number of vertices grows exponentially with the number of agents, making the approach applicable especially for problems with a few number of agents, and that is the case for car manufacturing stations where often two to four industrial robots are present. Another issue worth to mention is that a pure centralized path planning approach would require generating motions for a generalized robot whose number of degrees of freedom (dofs) is the sum of the dofs of each single robot. Considering that a typical station has often four robots and that each industrial robot usually has six dofs, then it is likely to think that iterative and decoupled approaches avoiding collisions are still motivated. The algorithm described above, indeed, requires path planning for each robot alone, not considering the others.

In Fig. \ref{fig_20kroA100_before_swept} three tours are illustrated after running the B\&B algorithm and in Fig. \ref{fig_20kroA100_after_swept} the tours after the synchronization.

The GTSP solver used is not an exact one, but it has good performance on the SGTSP instances from TSPLIB, see \cite{EKST}. It computes a starting tour and tries to improve it, by some local search techniques: some of them are similar to the ones described in \cite{KAR}. Anyway, any other GTSP solver algorithm may be used, among the many present in the literature, see \cite{KAR}. Near optimal solvers, especially if proved to be very powerful, may also lead to high quality global solutions.

\subsection{Asynchronous smoothing} 
\label{sec_asynchronous}

The results obtained through the synchronous routing may be further reduced when dropping the "synchronous" assumption. By doing that, the agents are allowed to move asynchronously while keeping the constraint of never being simultaneously on conflicting arcs and keeping the ordered sequence of vertices that have been assigned. The problem can therefore be modeled with a classical MILP formulation:
\\
\begin{subequations} 
\label{eq_MILP_COORDINATION}
\text{minimize} $c$ \text{s.t.}
\begin{align}
c \geq & c(T_k) & & \forall k=1,\ldots,N_A \label{eq_MILP_COORDINATION_1}\\
c(T_k) \geq & t[s^k_{N_{T_k}}]^L + c (s^k_{N_{T_k}}, s^k_1) & & \forall k=1,\ldots,N_A \label{eq_MILP_COORDINATION_2}\\
t[s^k_1]^L \geq & 0 & & \forall k=1,\ldots,N_A \label{eq_MILP_COORDINATION_INIT}\\
t[s^k_{i+1}]^A \geq & t[s^k_i]^L + c (s^k_i, s^k_{i+1}) & & \forall i=1,\ldots,N_{T_k}-1 \nonumber\\
t[s^k_{i+1}]^L \geq & t[s^k_{i+1}]^A & & \forall i=1,\ldots,N_{T_k}-1 \nonumber\\
& & & \forall k=1,\ldots,N_A \label{eq_MILP_COORDINATION_PRIO}\\
t[s^k_i]^L \geq & t[s^l_{j+1}]^A - Mb_z & & \label{eq_MILP_COORDINATION_MUTEX}\\
t[s^l_j]^L \geq & t[s^k_{i+1}]^A - M(1-b_z) & & \forall (s^k_i, s^k_{i+1}), (s^l_j, s^l_{j+1}) \nonumber\\
 & & & \text{ in \emph{conflict}} \nonumber\\
b_z \in & \{0,1\} & &
\end{align}
\end{subequations}

In this formulation there are priority constraints, see (\ref{eq_MILP_COORDINATION_2}), (\ref{eq_MILP_COORDINATION_INIT}), and (\ref{eq_MILP_COORDINATION_PRIO}), which simply state the order of points and their minimum timing schedule. The mutual exclusion constraints (\ref{eq_MILP_COORDINATION_MUTEX}) imply that no active conflict should be present in the optimal solution.

This problem can be solved by a MILP solver or by more specialized algorithms, as the one in \cite{SPENS_CASE}. As investigated in \cite{KOB}, instances of this problem involving only two agents can be efficiently solved in polynomial time by an A* search algorithm.

\begin{figure}
\centering
\includegraphics[width=2.5in]{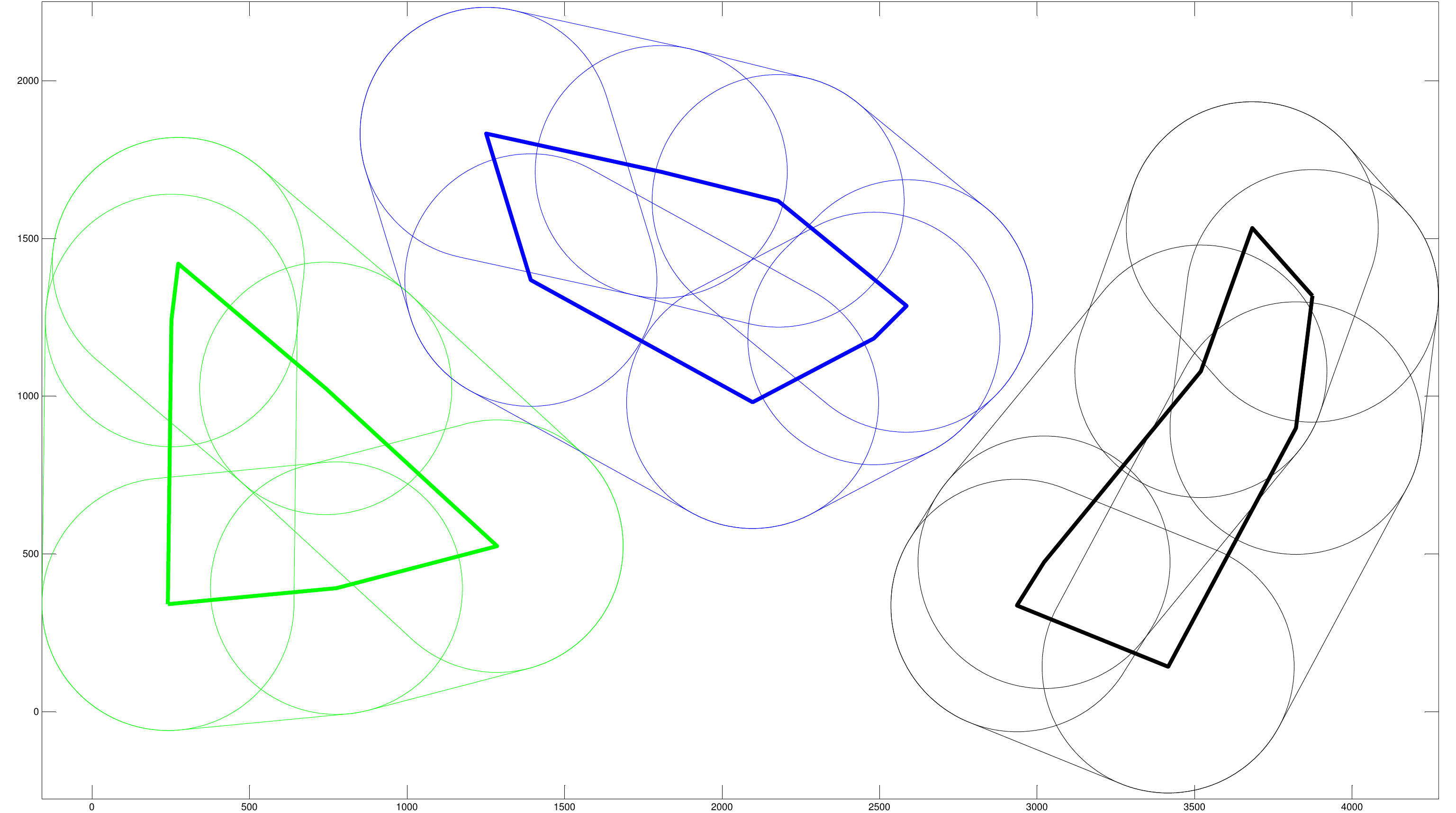}
\caption{Solution to instance 20kroA100 from TSPLIB, after B\&B: swept paths are drawn.}
\label{fig_20kroA100_before_swept}
\end{figure}

\begin{figure}
\centering
\includegraphics[width=2.5in]{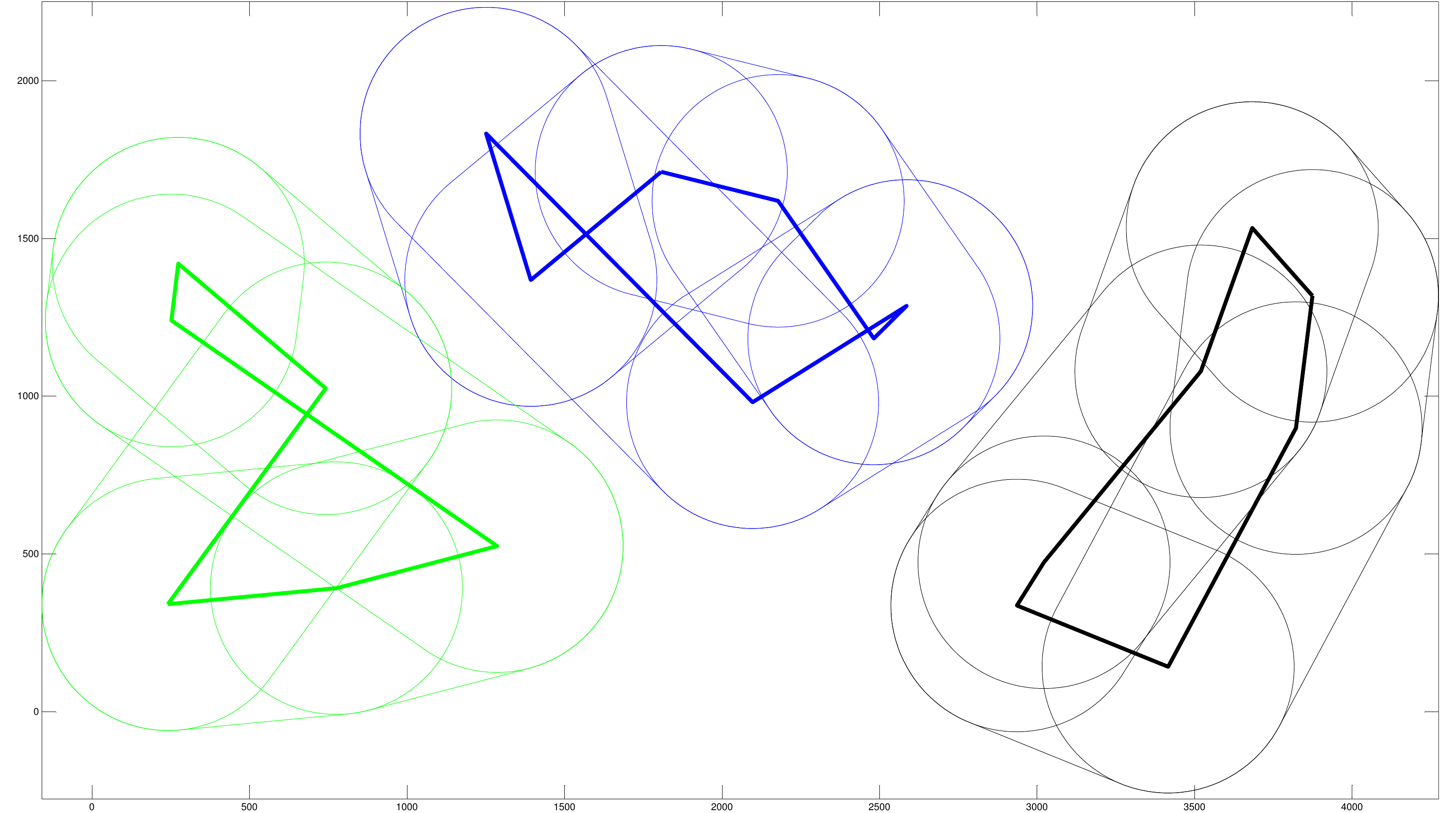}
\caption{Solution to instance 20kroA100 from TSPLIB, after synchronization: swept paths are drawn.}
\label{fig_20kroA100_after_swept}
\end{figure}

Note that, for the example illustrated in Fig. \ref{fig_rerouting}, with solution in Fig. \ref{fig_ReroutingStateSpace}, by letting the agents move asynchronously, the cycle time can be further reduced from 6 to 4, whereas the value obtained in the classical way can not be shortened at all, remaining 6.8. The test case illustrated in Fig. \ref{fig_20kroA100_before_swept} has a cycle time of 4273 when scheduling the tours according to formulation (\ref{eq_MILP_COORDINATION}). Note that, assuming the agents being circles, the area that each agent sweeps is illustrated and collisions between paths can be identified, see Fig. \ref{fig_20kroA100_before_swept} and Fig. \ref{fig_20kroA100_after_swept}. After applying the synchronous routing algorithm, the tours whose paths collide are changed such that the potential collision areas are decreased. This result may or may not lead to better cycle times. In this example, by maintaining these paths, dropping the synchronous hypothesis, and by coordinating the paths, the cycle time is improved to 3963, see  Fig. \ref{fig_20kroA100_after_swept}.

\section{Robot Path Planning}

So far, we have assumed that the robot paths durations have already been generated. We have neglected collisions between the robots and the environment, and computed the time $c(q_S, q_E)$ it takes for one robot to move from a start configuration $q_S$ to an end configuration $q_E$ as 

\begin{equation}
\label{MAX_NORM}
c(q_S, q_E) = \max \limits_{i=1,\ldots,N_J} \left\{ \frac{|q_S(i)-q_E(i)|}{\omega_{MAX}(i)} \right\}
\end{equation}

where, $q_S(i)$ and $q_E(i)$ are the $i$-th component for the start and end configuration, respectively, and $\omega_{MAX}(i)$ is the maximum angular velocity for the $i$-th robot joint. Note that often industrial robots have six rotational joints.

In reality, however, straight paths in the robot joint space may collide with the environment. This motivates the use of robot path planning techniques to provide the robots with point-to-point collision-free paths. Several approaches are suitable, see \cite{LATOMBE_BOOK} and \cite{LAVALLE_BOOK}, mostly based on sampling methods: two of these methods are Probabilistic Roadmap Method, see \cite{KAVRAKI} and \cite{BOBO}, and Rapidly-Exploring Random Trees (RRT), see \cite{KUFFNER}. The performance of these algorithms heavily rely on fast and accurate collision and distance computations, which nowadays involve triangle models and point cloud, see \cite{TAFURI, SHELL1, SHELL2}. Here, we have used the built-in robot path planning functionalities in IPS, see \cite{IPS-online}. 

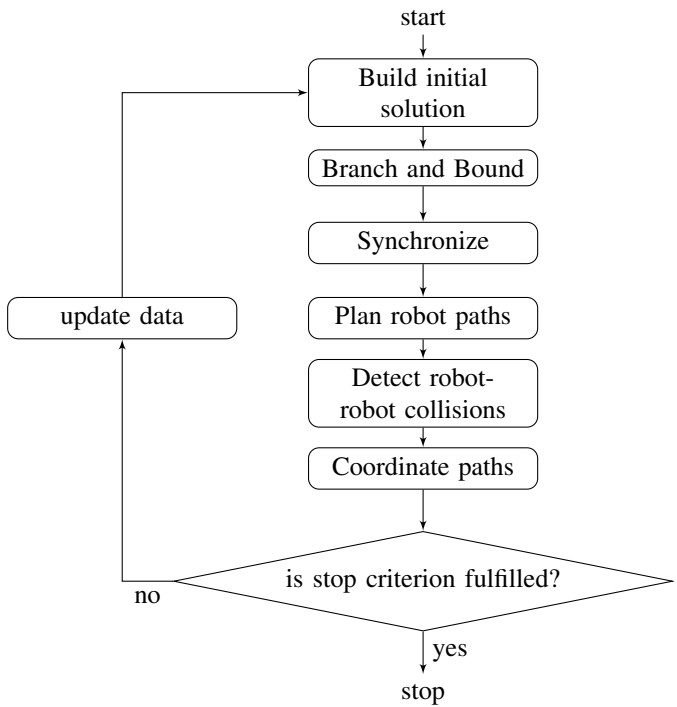
\begin{figure}
\usetikzlibrary{shapes,arrows}
\centering
%
\tikzstyle{decision} = [diamond, draw, aspect=5, text centered, node distance=1cm]
\tikzstyle{block} = [rectangle, draw, text width=8em, text centered, rounded corners]
\tikzstyle{line} = [draw, -latex']
\begin{tikzpicture}[node distance = 1cm, auto]
\node [block] (init) {Build initial solution};
\node [above of=init] (start) {start};
\node [block, below of=init] (BB) {Branch and Bound};
\node [block, below of=BB] (synch) {Synchronize};
\node [block, below of=synch] (PP) {Plan robot paths};
\node [block, left of=PP, node distance=4cm] (update) {update data};
\node [block, below of=PP] (collision) {Detect robot-robot collisions};
\node [block, below of=collision] (coordinate) {Coordinate paths};
\node [decision, below of=coordinate, node distance=1.5cm] (decide) {is stop criterion fulfilled?};
\node [below of=decide, node distance=1.5cm] (stop) {stop};
%
\path [line] (start) -- (init);
\path [line] (init) -- (BB);
\path [line] (BB) -- (synch);
\path [line] (synch) -- (PP);
\path [line] (PP) -- (collision);
\path [line] (collision) -- (coordinate);
\path [line] (coordinate) -- (decide);
\path [line] (decide) -| node [near start] {no} (update);
\path [line] (update) |- (init);
\path [line] (decide) -- node {yes}(stop);
\end{tikzpicture}

\caption{Iterative lazy optimization.}
\label{fig_flowchart}
\end{figure}

Since collision handling may be very computationally expensive, we adopt here a lazy strategy, \cite{BOBO}. By "lazy" it is meant that computations known to be demanding in terms of time and space complexity are postponed, run after the faster parts of the algorithm. In our case, see Fig. \ref{fig_flowchart}, the computation of collision-free paths, the detection of robot-robot collisions and the paths coordination are the last steps of the iterative procedure. This strategy is based on an optimistic view of the problem: if there were no collisions at all, then solutions obtained after the branch and bound would be optimal. The "update data" step consists in maintaining updated costs for the paths computed, information about which paths are colliding with each other, and, where possible, make some logical implications about the problem in order to derive information about the state space not yet explored. The stop criteria usually adopted in these cases are: maximum available time or maximum number of iterations reached, or the gap between the solution after the "Coordinate paths" and the one after "Branch and Bound" under a defined threshold.

\section{Industrial test case}

\begin{figure}
\centering
\includegraphics[width=2.5in]{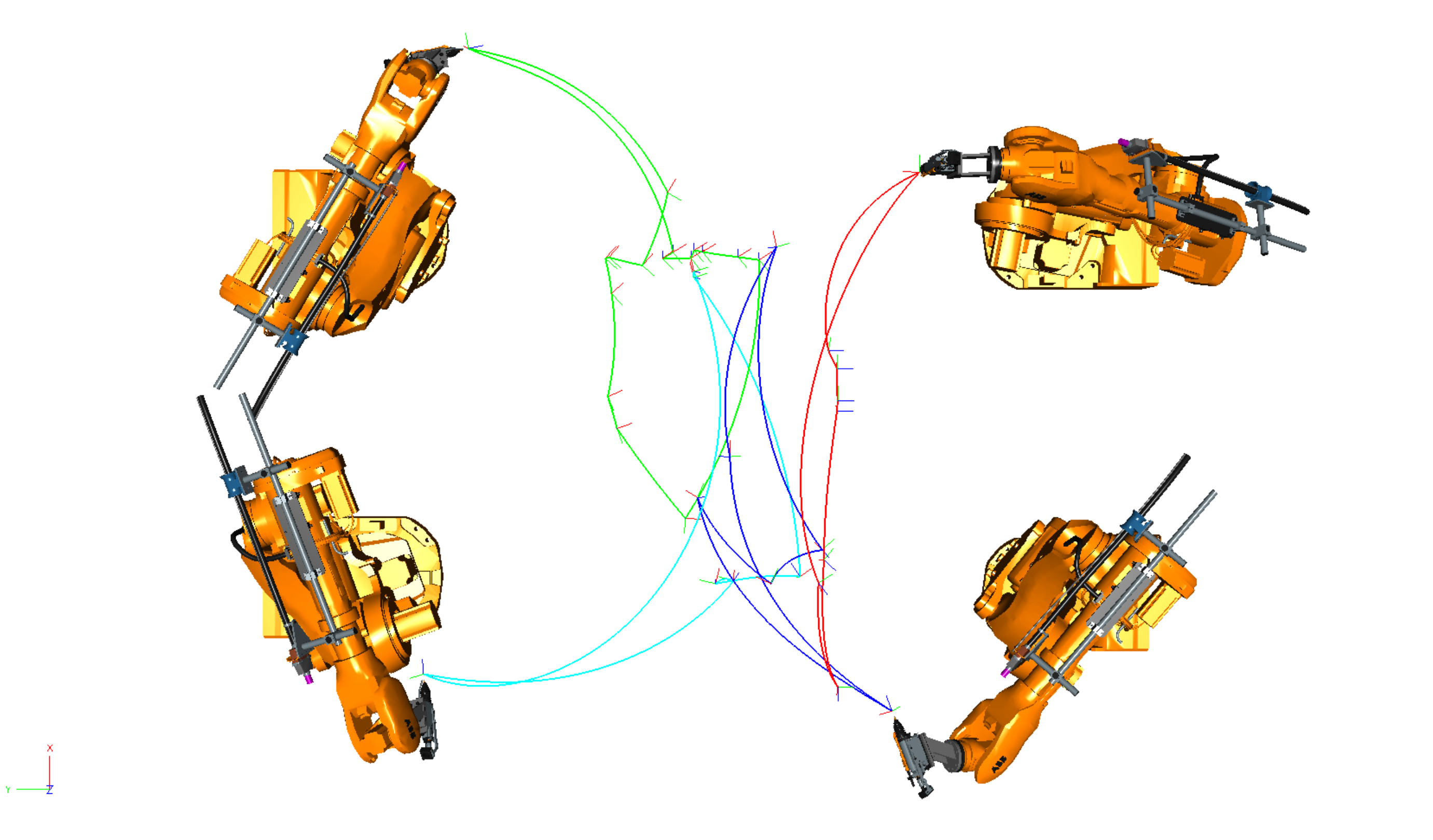}
\caption{Four robots with their home positions and 32 stud weld tasks: initial solutions.}
\label{fig_4coordNOTSYNCHED}
\end{figure}

\begin{figure}
\centering
\includegraphics[width=2.5in]{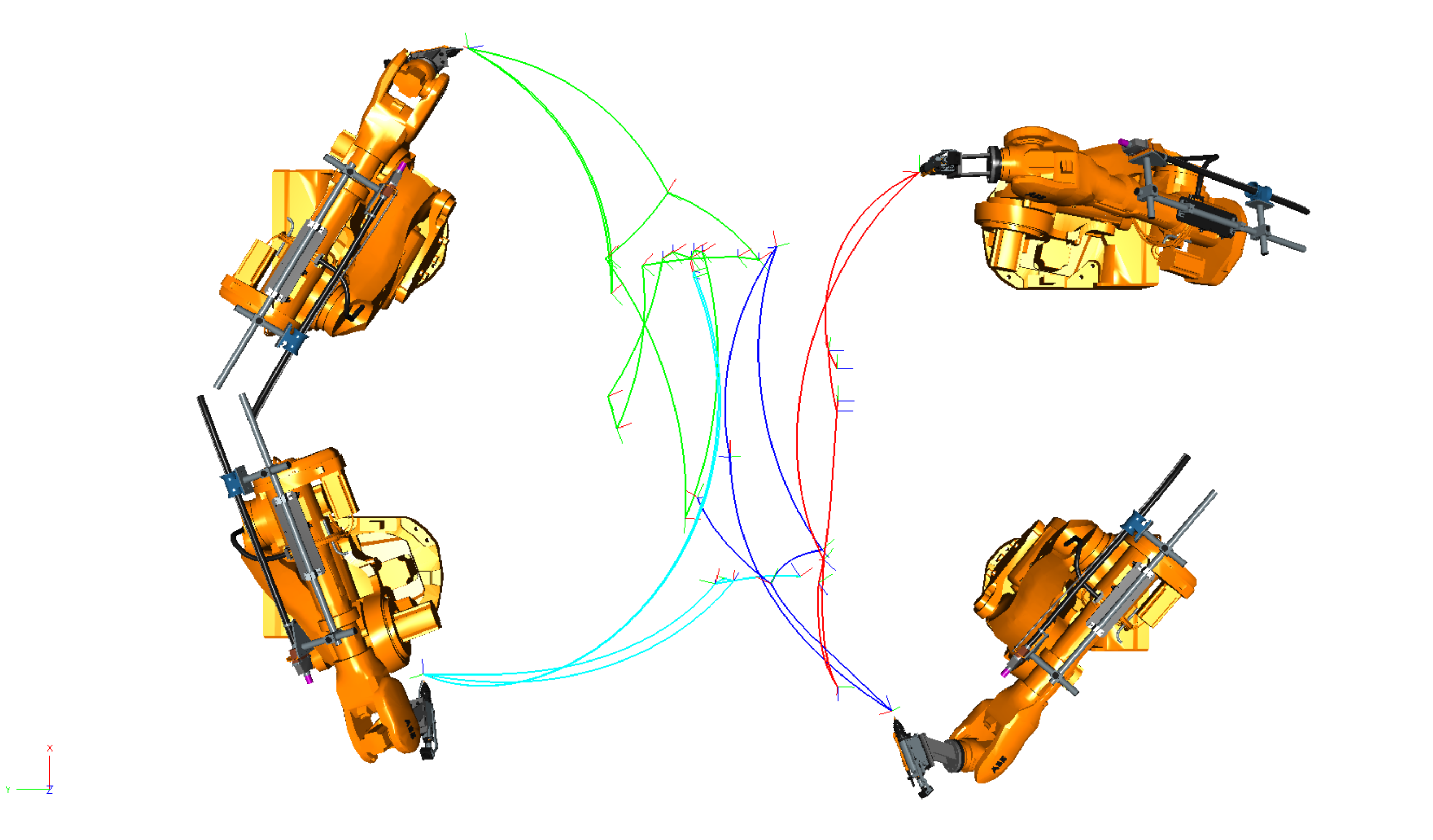}
\caption{Four robots with their home positions and 32 stud weld tasks: solutions after being synchronized.}
\label{fig_4coordSYNCHED}
\end{figure}

These algorithms have been interfaced and tested towards the simulation software IPS, \cite{IPS-online}, in order to handle more realistic problem instances. Here, we show the results obtained from the study of an industrial test case: robot models and welding points are courtesy of a major Swedish automotive manufacturer.
This industrial case consists of a stud welding station with four robots, their 4 home positions and 32 stud points. The first solution is obtained neglecting collisions between robots, and afterwards coordinating the paths obtained. The second solution, instead, allows re-sequencing the tasks within each robot, by applying the algorithm in Section \ref{sec_SRS} and is smoothed by coordinating the new paths obtained, see Section \ref{sec_asynchronous}. The cycle time is improved from 8.14s to 6.66s, thus a circa 18\% improvement. It is important to note that, in this case, the cycle time is not improved when considering the cost measure in (\ref{L_INF_NORM}), but only after the synchronous hypothesis is dropped, see Section \ref{sec_asynchronous}. By looking at Fig. \ref{fig_4coordSYNCHED}, it is possible to note how the cyan path and the blue ones are shifted in such a way that smaller collision areas are present, w.r.t. Fig. \ref{fig_4coordNOTSYNCHED}.

\begin{figure}
\centering
\includegraphics[width=2.5in]{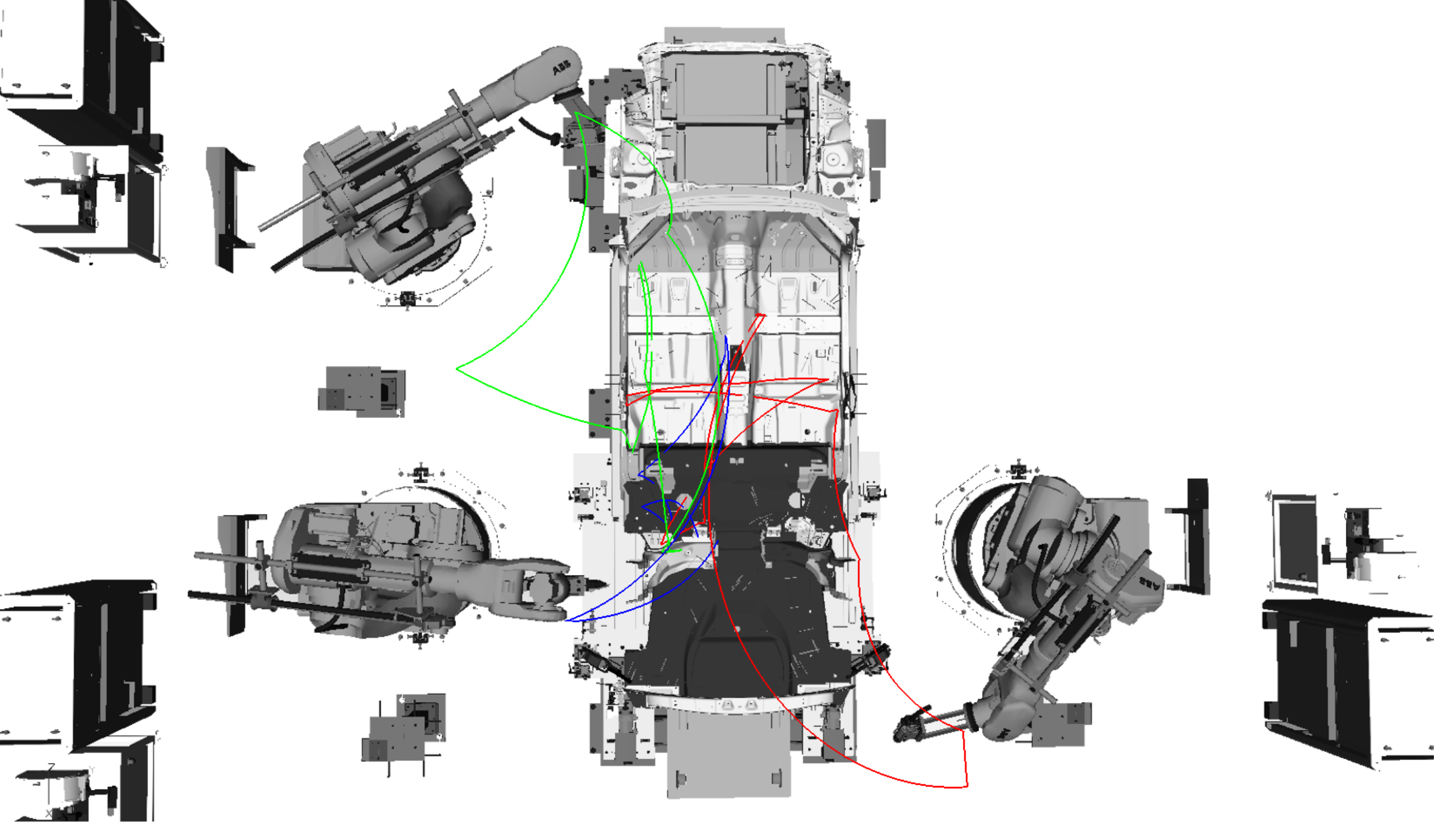}
\caption{Three robots with their home positions and 20 stud weld tasks: solutions after being synchronized. Courtesy of Volvo Cars.}
\label{fig_3coordSYNCHED}
\end{figure}

Another test case adapted from a stud welding station has been successfully solved. It consists of 3 robots sharing 20 tasks. The final paths are illustrated in Fig. \ref{fig_3coordSYNCHED}, where their respective TCP traces are drawn. The robots will move in a collision free way among the assigned tasks avoiding collisions with the environment and among each others, with the aim to minimize the makespan. These solutions were obtained in less than ten iterations. 

\section{Conclusion}

In this work we have studied the problem of generating, routing and scheduling robot motions in car manufacturing stations. We have proposed an iterative decoupled approach, considering scenarios without robot collisions, and scenarios for highly cluttered environments. An exact solution for the problem without collisions is obtained by a B\&B algorithm that does not require LP based tools and is very easy to implement. The lower bounds used are a generalization of previous approaches present in the literature.

The novel algorithm to resolve conflicts between robot paths has shown good results when solving high cluttered environments problems, giving the possibility to modify already computed motions to avoid collisions and even to improve cycle time.

The possibility to interact with Computer Aided Manufacturing software and, therefore, to generate robot programs has been demonstrated on industrial test cases.

Good results have been shown for the optimization of stations with up to 40 tasks, 4 robots and tens of possibilities to perform each task, \emph{e.g.} geometry stations. However, efficient heuristics are needed to handle other types of stations with more than 100 tasks, \emph{e.g.} assembly stations.

Several questions can be further investigated. Among them, a less decoupled B\&B where at each leaf the synchronous routing and the asynchronous smoothing are solved.


\section*{Acknowledgement}

The authors would like to thank the entire personnel at the Geometry and Motion Planning Group at the Fraunhofer-Chalmers Research Centre for Industrial Mathematics, for useful discussions and support in the implementation, in particular Daniel Segerdahl. The authors also thank the anonymous referees and associate editor for their helpful suggestions.


\ifCLASSOPTIONcaptionsoff
  \newpage
\fi




\appendices
\section{}
\label{APPENDIX}

\newtheorem{theorem_proof}{Lemma}
\begin{theorem_proof}
Given a best tour $T^*_n$ with length $c(T^*_n)$ for an SGTSP $G_n$ with $n$ groups, and an SGTSP $G_{n+1} = G_n \cup \{g_{n+1}\}$, if the triangle inequality holds, then
\begin{equation} 
\label{theo_TRIANGLE_INEQ_APP}
c(T^*_{n+1}) \geq \max \{c(T^*_n) + 2 \min (g_{n+1}, G_n) - \max (G_n), c(T^*_n)\}
\end{equation}
where $c(T^*_{n+1})$ is the length for an optimal tour in $G_{n+1}$, $\min (g_{n+1}, G_n) = \min \limits_{\substack{p_i \in G_n \\ p_k \in g_{n+1}}} c(p_i, p_k)$, and $\max (G_n) = \max \limits_{\substack{p_i \neq p_j \in G_n}} c(p_i, p_j)$.
\end{theorem_proof}

\begin{IEEEproof}
Suppose we are given a best tour $T^*_{n+1}= (p^*_1, \ldots, p^*_{n}, p^*_{n+1}, p^*_1)$, with length $c(T^*_{n+1})$, in $G_{n+1}$. A feasible tour in $G_n$ is $T_n = (p^*_1, \ldots, p^*_{n}, p^*_1)$, which has a total length of 

\begin{equation} 
\label{eq_FEASIBLE_TOUR}
c(T_n) = c(T^*_{n+1}) - c(p^*_{n}, p^*_{n+1}) - c(p^*_{n+1}, p^*_1) + c (p^*_{n}, p^*_1)
\end{equation}

A best tour $T^*_n$, thus, satisfies:

\begin{equation} 
\label{proof_TRIANGLE_INEQ_1}
c(T^*_n) \leq c(T^*_{n+1}) - c(p^*_{n}, p^*_{n+1}) - c(p^*_{n+1}, p^*_1) + c (p^*_{n}, p^*_1)
\end{equation}

We have, therefore, a lower bound for $T^*_{n+1}$. However, this bound uses information about a best tour in $G_{n+1}$, 
which we do not want to compute. In order to eliminate such dependency, one can use the fact that $c(p^*_{n}, p^*_{n+1}) \geq \min (g_{n+1}, G_n)$, $c(p^*_{n+1}, p^*_1) \geq \min (g_{n+1}, G_n)$, and $c (p^*_{n}, p^*_1) \leq \max (G_n)$. These yield to

\begin{equation} 
\label{proof_TRIANGLE_INEQ_2}
c(T^*_{n+1}) \geq c(T^*_n) + 2\min (g_{n+1}, G_n) - \max (G_n)
\end{equation}

The first part of the lemma has been proved without using the triangle inequality. For the second part, by exploiting the triangle inequality among $p^*_n$, $p^*_{n+1}$, and $p^*_1$, we have, from (\ref{eq_FEASIBLE_TOUR}):

\begin{equation} 
\label{proof_TRIANGLE_INEQ_3}
c(T^*_{n+1}) - c(T_n) = c(p^*_{n}, p^*_{n+1}) + c(p^*_{n+1}, p^*_1) - c (p^*_{n}, p^*_1) \geq 0
\end{equation}

It follows directly:

\begin{equation} 
\label{proof_TRIANGLE_INEQ_4}
c(T^*_{n+1}) \geq c(T_n) \geq c(T^*_n)
\end{equation}

\end{IEEEproof}

\begin{theorem_proof}
\label{theo_INTEGER_RELAXATION}
The optimum $d^* $ for the following problem

\begin{subequations} 
\label{GREEDY_SET_PART_d}
\text{minimize} $d$ \text{s.t.}
\begin{align}
d - \sum\limits_{i:g_i \in G_N} x_{ik} &\ge d_k & \forall k=1,\ldots,N_A \label{GREEDY_SET_PART_1_d}\\
\sum\limits_{k=1}^{N_A} x_{ik} &= 1 & \forall i:g_i \in G_N \label{GREEDY_SET_PART_2_d}\\
x_{ik} \in &\{0,1\} & \forall i:g_i \in G_N \nonumber\\
 & & \forall k=1,\ldots,N_A \nonumber
\end{align}
\end{subequations}

is given by

\begin{equation} 
\label{GREEDY_SET_PART_INTEGER_d}
d^* = \begin{cases}
d_{\overline{k}} & \text{if $|G_N| \leq \Delta N$}\\
\min \limits_{j=1,\ldots,N_A} \left\{ d_j + \left \lceil {\frac{|G_N| + \sum\limits_{k=1}^{N_A} \lceil d_k - d_j \rceil}{N_A}} \right \rceil \right\} & \text{otherwise}
\end{cases}
\end{equation}

where $d_{\overline{k}} = \max \limits_{k=1,\ldots,N_A} \{d_k\}$ and $\Delta N = \sum\limits_{i = 1}^{N_A} \lfloor d_{\overline{k}} - d_i \rfloor$.

\end{theorem_proof}

\section{}
\label{APPENDIX_2}

\begin{figure}
\centering
\def\svgwidth{2 in}
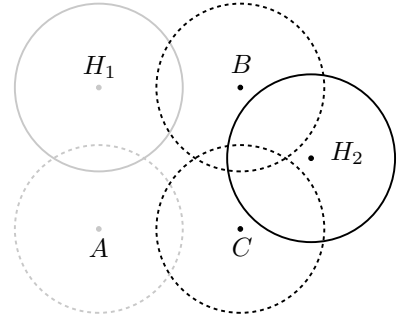
\caption{An example with two agents where there exists a feasible solution, but by the transformation to a GTSP it is not possible to find any.}
\label{fig_Deadlock}
\end{figure}

Some problems remain still unsolvable, even if a feasible solution exists, see Fig. \ref{fig_Deadlock}. A feasible solution is, \emph{e.g.}, $(H_1, H_2)$, $(A, H_2)$, $(A, B)$, $(A, H_2)$, $(H_1, H_2)$, $(H_1, C)$, $(H_1, H_2)$, whereas by the transformation to a GTSP it is not possible to find any.

\end{document}